\documentclass[10pt,twocolumn,letterpaper]{article}

\usepackage{color}
\usepackage{bm}
\usepackage[usenames,dvipsnames,svgnames,table]{xcolor}
\usepackage{caption}
\usepackage{siunitx} 
\usepackage{adjustbox}
\usepackage{multirow}
\usepackage{listings}
\usepackage{minibox}
\usepackage{xspace}
\usepackage{threeparttable}

\makeatletter
\DeclareRobustCommand\onedot{\futurelet\@let@token\@onedot}
\def\@onedot{\ifx\@let@token.\else.\null\fi\xspace}

\def\etal{\emph{et al}\onedot}

\makeatother

\usepackage[T1]{fontenc}
\usepackage[font=small,labelfont=bf,tableposition=top]{caption}
\DeclareCaptionLabelFormat{andtable}{#1~#2  \&  \tablename~\thetable}

\newcommand{\Tref}[1]{Table~\ref{#1}}
\newcommand{\tref}[1]{Table~\ref{#1}}
\newcommand{\eref}[1]{Equation~(\ref{#1})}

\newcommand{\fref}[1]{Figure~\ref{#1}}

\newcommand{\Fref}[1]{Figure~\ref{#1}}

\newcommand{\rot}{\mathbf{R}}
\newcommand{\trans}{\mathbf{t}}
\newcommand{\extrinsics}{\left[~\rot~|~\trans~\right]}

\newcommand{\vuh}{\tilde{\mathbf{u}}}
\newcommand{\vph}{\tilde{\mathbf{p}}}

\newcounter{todos}

\makeatletter
\newcommand{\figcaption}[1]{\def\@captype{figure}\caption{#1}}
\newcommand{\tblcaption}[1]{\def\@captype{table}\caption{#1}}
\makeatother

\usepackage[pagenumbers]{cvpr} 

\usepackage{graphicx}
\usepackage{amsmath}
\usepackage{amssymb}
\usepackage{booktabs}

\usepackage[pagebackref,breaklinks,colorlinks]{hyperref}

\usepackage[capitalize]{cleveref}
\crefname{section}{Sec.}{Secs.}
\Crefname{section}{Section}{Sections}
\Crefname{table}{Table}{Tables}
\crefname{table}{Tab.}{Tabs.}

\def\cvprPaperID{5864} 
\def\confName{CVPR}
\def\confYear{2022}

\begin{document}

\title{Rethinking Generic Camera Models for Deep Single Image Camera Calibration to Recover Rotation and Fisheye Distortion}

\author{
Nobuhiko Wakai$^1$ \quad\quad Satoshi Sato$^1$ \quad\quad Yasunori Ishii$^1$ \quad\quad Takayoshi Yamashita$^2$\\
$^1$ Panasonic Corporation\quad\quad $^2$ Chubu University\\
{\tt\small $\{lastname.firstname\}$@jp.panasonic.com} \quad\quad {\tt\small takayoshi@isc.chubu.ac.jp}}

\maketitle

\begin{abstract}
Although recent learning-based calibration methods can predict extrinsic and intrinsic camera parameters from a single image, the accuracy of these methods is degraded in fisheye images. This degradation is caused by mismatching between the actual projection and expected projection. To address this problem, we propose a generic camera model that has the potential to address various types of distortion. Our generic camera model is utilized for learning-based methods through a closed-form numerical calculation of the camera projection. Simultaneously to recover rotation and fisheye distortion, we propose a learning-based calibration method that uses the camera model. Furthermore, we propose a loss function that alleviates the bias of the magnitude of errors for four extrinsic and intrinsic camera parameters. Extensive experiments demonstrated that our proposed method outperformed conventional methods on two large-scale datasets and images captured by off-the-shelf fisheye cameras. Moreover, we are the first researchers to analyze the performance of learning-based methods using various types of projection for off-the-shelf cameras.
\end{abstract}

\section{Introduction}
\begin{figure}[t]
\centering
\includegraphics[width=1.00\hsize]{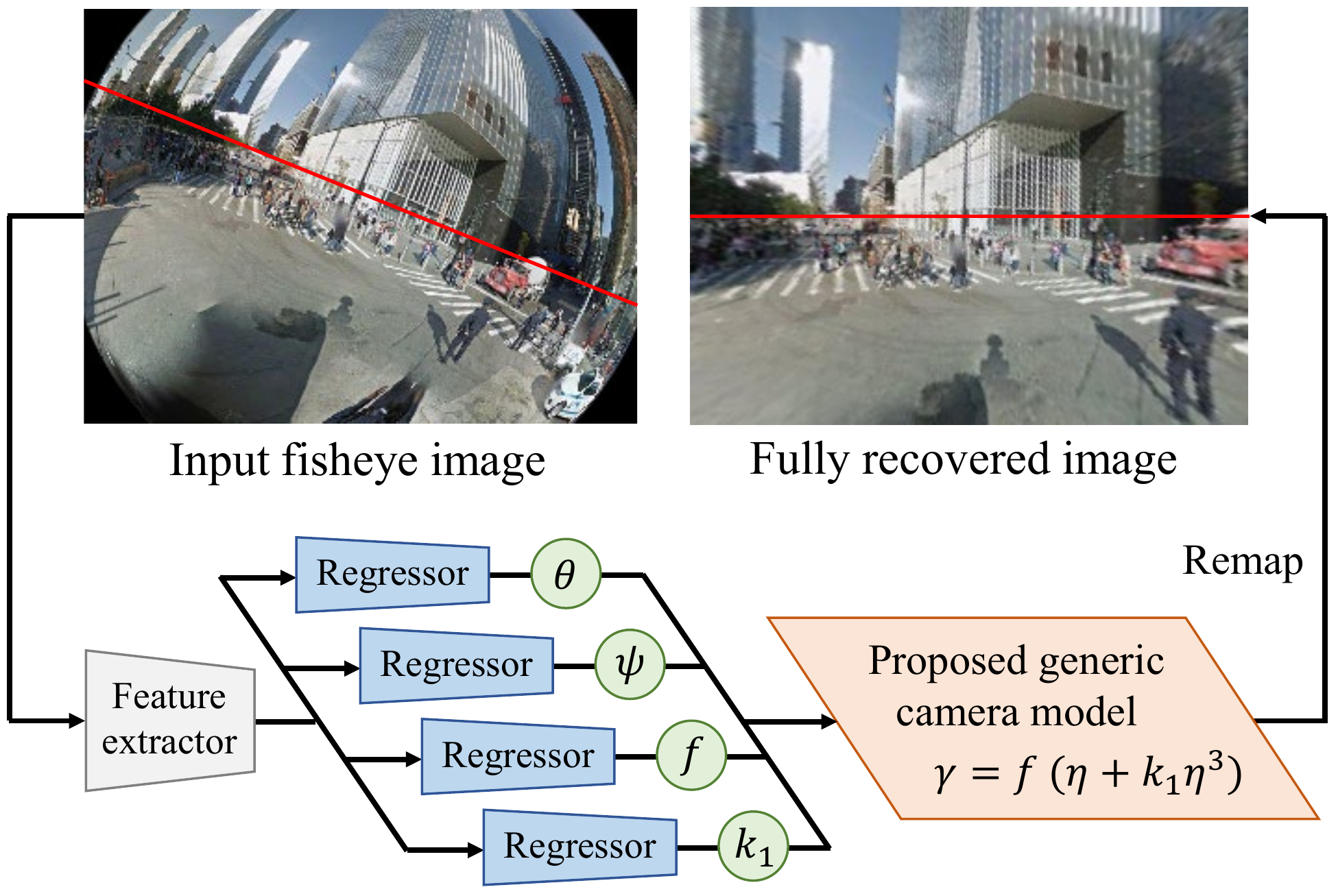}
\caption{Concept illustrations of our work. Our network predicts parameters in our proposed generic camera model to obtain fully recovered images using remapping. Red lines indicate horizontal lines in each of the images.}
\label{fig:concept}
\vspace{-1mm}
\end{figure}
Learning-based perception methods are widely used for surveillance, cars, drones, and robots. These methods are well established for many computer vision tasks. Most computer vision tasks require undistorted images; however, fisheye images have the superiority of a large field of view in visual surveillance~\cite{Fu2021}, object detection~\cite{Redmon2016}, pose estimation~\cite{Cao2017}, and semantic segmentation~\cite{LiuC2019}. To use fisheye cameras through removing distortion, camera calibration is a desirable step before perception. Camera calibration is a long-studied topic in areas of computer vision, such as image undistortion~\cite{liao2020,Yin2018}, image remapping~\cite{Wakai2021}, virtual object insertion~\cite{Hold-Geoffroy2018}, augmented reality~\cite{Alhaija2018}, and stereo measurement~\cite{Locher2016}. In camera calibration, we cannot escape the trade-off between accuracy and usability that we need a calibration object; hence, tackling the trade-off has been an open challenge, which we explain further in the following.

Calibration methods are classified into two categories: geometric-based and learning-based methods. Geometric-based calibration methods achieve high accuracy, but they require a calibration object, such as a cube\cite{Tsai1987} and planes~\cite{Zhang2000}, to obtain a strong geometric constraint. By contrast, learning-based methods can calibrate cameras without a calibration object from a general scene image~\cite{liao2020,Yin2018}, which is called deep single image camera calibration. Although learning-based methods do not require a calibration object, the accuracy of these methods is degraded for fisheye images because of the mismatch between the actual projection and expected projection in conventional methods. In particular, calibration methods~\cite{Lopez2019,Wakai2021} that predict both camera rotation and distortion have much room for improvement regarding addressing complex fisheye distortion. L\'{o}pez-Antequera's method~\cite{Lopez2019} was designed for non-fisheye cameras with radial distortion and cannot process fisheye distortion. Although four standard camera models are used for fisheye cameras, Wakai's method~\cite{Wakai2021} supports only one fisheye camera model.

Based on the observations above, we propose a new generic camera model for various fisheye cameras. The proposed generic camera model has the potential to address various types of distortion. For the generic camera model, we propose a learning-based calibration method that predicts extrinsic parameters (tilt and roll angles), focal length, and a distortion coefficient simultaneously from a single image, as shown in~\fref{fig:concept}. Our camera model is utilized for learning-based methods through a closed-form numerical calculation of camera projection. To improve the prediction accuracy, we use a joint loss function composed of each loss for the four camera parameters. Unlike heuristic approaches in conventional methods, our loss function makes significant progress; that is, we can determine the optimal joint weights based on the magnitude of errors for these camera parameters instead of the heuristic approaches.

To evaluate the proposed method, we conducted extensive experiments on two large-scale datasets~\cite{Mirowski2019,Chang2018} and images captured by off-the-shelf fisheye cameras. This evaluation demonstrated that our method meaningfully outperformed conventional geometric-based~\cite{Santana2016} and learning-based methods~\cite{Chao2020,liao2020,Lopez2019,Yin2018,Wakai2021}. The major contributions of our study are summarized as follows:
\begin{itemize}
\item We propose a learning-based calibration method for recovering camera rotation and fisheye distortion using the proposed generic camera model that has an adaptive ability for off-the-shelf fisheye cameras. To the best of our knowledge, we are the first researchers to calibrate extrinsic and intrinsic parameters of generic camera models from a single image.

\item We propose a new loss function that alleviates the bias of the magnitude of errors between the ground-truth and predicted camera parameters for four extrinsic and intrinsic parameters to obtain accurate camera parameters.

\item We first analyze the performance of learning-based methods using various off-the-shelf fisheye cameras. In previous studies, these conventional learning-based methods were evaluated using only synthetic images.
\end{itemize}

\section{Related work}
\label{sec:related_work}
\textbf{Camera calibration:} Camera calibration estimates parameters composed of extrinsic parameters (rotation and translation) and intrinsic parameters (image sensor and distortion parameters). Geometric-based calibration methods have been developed using a strong constraint based on the calibration object~\cite{Tsai1987,Zhang2000} or line detection~\cite{Aleman2014,Santana2016}. This constraint explicitly represents the relation between world coordinates and image coordinates for the stable optimization of calibration. By contrast, learning-based methods based on convolutional neural networks calibrate cameras from a single image in the wild. In this study, we focus on learning-based calibration methods and describe them below.

Calibration methods for only extrinsic parameters have been proposed that are aimed at narrow view cameras~\cite{Huang2019,Nie2020,Saputra2019,Sha2020,Xian2019,XueF2019} and panoramic $360^{\circ}$ images~\cite{Davidson2020}. These methods cannot calibrate intrinsic parameters, that is, they cannot remove distortion. For extrinsic parameters and focal length, narrow-view camera calibration was developed with depth estimation~\cite{Chen2019,Gordon2019} and room layout~\cite{Ren2020}. These methods are not suitable for fisheye cameras because fisheye distortion is not negligible because the projection of the field of view is over $180^{\circ}$.

To address large distortion, calibration methods for only undistortion have been proposed that use specific image features, that is, segmentation information~\cite{Yin2018}, straight lines~\cite{XueZ2019}, and ordinal distortion of part of the images~\cite{liao2020}. Furthermore, Chao~\etal{}~\cite{Chao2020} proposed undistortion networks based on generative adversarial networks~\cite{Goodfellow2014}. These methods can process only undistortion and image remapping tasks.

For both extrinsic and intrinsic parameters, L\'{o}pez-Antequera~\etal{}~\cite{Lopez2019} proposed a pioneering method for non-fisheye cameras. This method estimates distortion using a polynomial function model of perspective projection similar to Tsai's quartic polynomial model~\cite{Tsai1987}. This polynomial function of the distance from a principal point has the two coefficients for second- and fourth-order terms. The method is only trainable for the second-order coefficient, and the fourth-order coefficient is calculated using a quadratic function of the second-order one. This method does not calibrate fisheye cameras effectively because the camera model does not represent fisheye camera projection. Additionally, Wakai~\etal{}~\cite{Wakai2021} proposed a calibration method for extrinsic parameters and focal length in fisheye cameras. Although four types of standard fisheye projection are used for camera models, 
for example, equisolid angle projection, the method of Wakai~\etal{}~\cite{Wakai2021} only expects equisolid angle projection. As discussed above, conventional learning-based calibration methods do not fully calibrate extrinsic and intrinsic parameters of generic camera models from a single image.

\textbf{Exploring loss landscapes:} To optimize networks effectively, loss landscapes have been explored after training~\cite{Dauphin2018,Goodfellow2015,Li2018} and during training~\cite{Groenendijk2020}. In learning-based calibration methods, we have the problem that joint weights are difficult to determine without training. To stabilize training or the merging of heterogeneous loss components, the joint loss function was often defined~\cite{liao2020,Lopez2019,Wakai2021,Yin2018}. However, these joint weights were defined using experiments or the same values, that is, unweighted joints. These joint weights are hyperparameters that depend on networks and datasets. A hyperparameter search method was proposed by Akiba~\etal{}~\cite{Akiba2019}. However, hyperparameter search tools require large computational cost because they execute various conditions. Additionally, to analyze the optimizers, Goodfellow~\etal{}~\cite{Goodfellow2015} proposed an examination method for loss landscapes that use linear interpolation from the initial network weights to the final weights. To overcome the saddle points of loss landscapes, Dauphin~\etal{}~\cite{Dauphin2018} proposed an optimization method based on Newton's method. Furthermore, Li~\etal{}~\cite{Li2018} developed a method for visualizing loss landscapes. Although these methods can explore high-order loss landscapes, the optimal values of joint loss weights have not been determined in learning-based calibration methods. Moreover, the aforementioned methods cannot explore loss landscapes without training because they require training results.

\section{Proposed method}
\label{sec:proposed_method}
First, we describe our proposed camera model based on a closed-form solution for various fisheye cameras. Second, we describe our learning-based calibration method for recovering rotation and fisheye distortion. Finally, we explain a new loss function, with its notation and mechanism.

\subsection{Generic camera model}
\label{subsec:camera_model}
Camera models are composed of extrinsic parameters $\extrinsics$ and intrinsic parameters, and these camera models represent the mapping from world coordinates $\vph$ to image coordinates $\vuh$ in homogeneous coordinates. This projection can be expressed for radial distortion~\cite{Puskorius1988} and fisheye models~\cite{Shah1994} as
\begin{align}\label{eq:nonlinear_model}
\vuh = 
\left[
\begin{array}{ccc}
\gamma / d_u & 0 & c_u \\
0 & \gamma / d_v & c_v \\
0 & 0 & 1
\end{array}
\right] \extrinsics \vph,
\end{align}
where $\gamma$ is distortion, $\left(d_u, d_v\right)$ is the image sensor pitch,  $\left(c_u, c_v\right)$ is a principal point, $\rot$ is a rotation matrix, and $\trans$ is a translation vector. The subscripts of $u$ and $v$ denote the horizontal and vertical direction, respectively.

The generic camera model including fisheye lenses~\cite{Gennery2006} is defined as 
\begin{equation}
\label{eq:generic_camera}
\gamma = \tilde{k_1} \eta + \tilde{k_2} \eta ^3 + \cdots,
\end{equation}
where $\tilde{k_1}$, $\tilde{k_2}, \ldots$ are distortion coefficients, and $\eta$ is an incident angle. Note that the focal length is not defined explicitly, that is, the focal length is set to~$1$ mm, and the distortion coefficients represent distortion and implicit focal length.

\subsection{Proposed camera model}
A generic camera model with high order has the potential to achieve high calibration accuracy. However, this high-order function leads to unstable optimization, particularly for learning-based methods. Considering this problem, we propose a generic camera model for learning-based fisheye calibration using the explicit focal length, given by
\begin{equation}
\label{eq:our_generic_camera}
\gamma = f(\eta + k_1 \eta ^3),
\end{equation}
where $f$ is the focal length and $k_1$ is a distortion coefficient.

\textbf{Evaluating our camera model:}
\setlength{\tabcolsep}{4pt}
\begin{table}[t]
\caption{Comparison of absolute errors in fisheye camera models}
\label{table:comparison_of_model}
\vspace{-1mm}
\centering
\scalebox{0.85}{
\begin{tabular}{ccccc}
\hline\noalign{\smallskip}
\multirow{2}{*}{Reference model$^1$} & \multicolumn{4}{c}{Mean absolute error [pixel]} \\
 & \phantom{0}STG\phantom{0} & \phantom{0}EQD\phantom{0} & \phantom{0}ESA\phantom{0} & \phantom{00}ORT\phantom{0} \\
\noalign{\smallskip}
\hline
\noalign{\smallskip}
Stereographic (STG) & -- & $\phantom{0}9.33$ & $13.12$ & $93.75$ \\
Equidistance (EQD) & $\phantom{0}9.33$ & -- & $\phantom{0}3.79$ & $23.58$  \\
Equisolid angle (ESA) & $13.12$ & $\phantom{0}3.79$ & -- & $14.25$ \\
Orthogonal (ORT) & $93.75$ & $23.58$ & $14.25$ & -- \\
\phantom{00}Proposed generic model\phantom{00} & \textbf{\phantom{0}0.54} & \textbf{\phantom{0}0.00} & \textbf{\phantom{0}0.02} & \textbf{\phantom{0}0.35} \\
\hline
\noalign{\smallskip}
\multicolumn{5}{l}{\scalebox{0.85}{~$^1$ Each reference model is compared with other fisheye models}} \\
\end{tabular}
}
\end{table}
\setlength{\tabcolsep}{1.4pt}
Our generic camera model is a third-order polynomial function that corresponds to Taylor's expansion of the trigonometric function in fisheye cameras, that is, stereographic projection, equidistance projection, equisolid angle projection, and orthogonal projection. In the following, we show that our model can express trigonometric function models with slight errors.

We compared projection function, $\gamma = g(\eta)$, of the four trigonometric function models and our generic camera model, as shown in~\tref{table:comparison_of_model}. In this comparison, we calculated mean absolute errors $\epsilon$ between pairs of the projection function $g_1$ and $g_2$. We defined the errors as $\epsilon = 1/(\pi/2) \int_0^{\pi/2} |~g_1(\eta) - g_2(\eta)~|~d\eta$. This mean absolute errors simply represents mean distance errors in image coordinates. Our model is useful for various fisheye models because our model had the smallest mean absolute errors among the camera models in~\tref{table:comparison_of_model}.

\textbf{Calculation easiness:}
For our generic camera model, it is easy to calculate back-projection, which converts image coordinates to corresponding incident angles. When using back-projection, we must solve the generic camera model against incident angles $\eta$ in~\eref{eq:our_generic_camera}. Practically, we can solve equations on the basis of iterations or closed-form. Non-fisheye cameras often use the iteration approaches~\cite{Alvarez2009}. By contrast, we cannot use the iteration approaches for fisheye cameras because large distortion prevents us from obtaining the initial values close to solutions. We therefore use a closed-form approach because the Abel-Ruffini theorem~\cite{Zoladek2000} shows that fourth-order or less algebraic equations are solvable.

\subsection{Proposed calibration method}
\label{subsec:proposed_calibration_method}
To calibrate various fisheye cameras, we propose a learning-based calibration method that uses our generic camera model. We use DenseNet-161~\cite{Huang2017} pretrained on ImageNet~\cite{FeiFei2015} to extract image features and details as follows: First, we convert the image features using global average pooling~\cite{lin2014} for regressors. Second, four individual regressors predict the normalized parameters (from $0$ to $1$) of the tilt angle $\theta$, roll angle $\psi$, focal length $f$, and a distortion coefficient $k_1$. Each regressor consists of a $2208$-channel fully connected (FC) layer with Mish activation~\cite{Misra2020} and a $256$-channel FC layer with sigmoid activation. Batch normalization~\cite{Ioffe2015} uses these FC layers. Finally, we predict a camera model by recovering the ranges of the normalized camera parameters to their original ranges. We scale the input images to $224 \times 224$ pixels following conventional studies~\cite{Hold-Geoffroy2018,Lopez2019,Wakai2021}.

\subsection{Harmonic non-grid bearing loss}
\label{subsec:our_loss}
\begin{figure}[t]
\centering
\includegraphics[width=1.00\hsize]{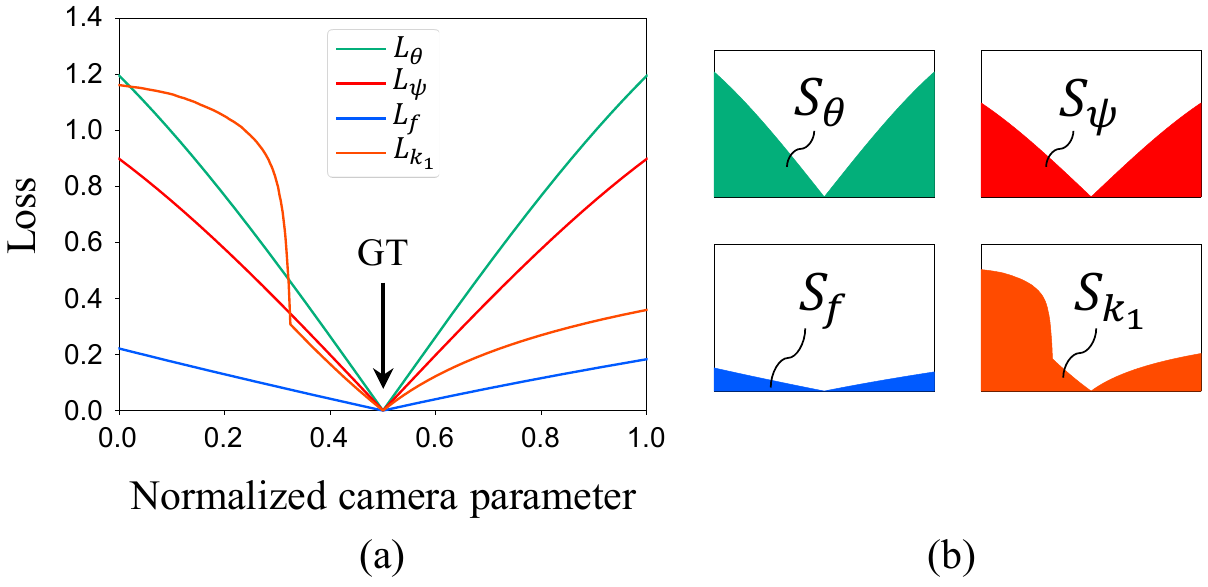}
\vspace{-6mm}
\caption{Difference between the non-grid bearing loss functions~\cite{Wakai2021} for the camera parameters. (a) Each loss landscape along the normalized camera parameters using a predicted camera parameter with a subscript parameter and ground-truth parameters for the remaining parameters, and the ground-truth values are set to $0.5$. (b) Areas $S$ integrating $L$ that shows (a) for the interval $0$ to $1$ for $\theta$, $\psi$, $f$, and $k_1$.}
\label{fig:harmonic_loss_plot}
\end{figure}
Unlike a loss function based on image reconstruction, Wakai~\etal{} proposed the non-grid bearing loss function~$L$~\cite{Wakai2021} based on projecting image coordinates to world coordinates as
\begin{eqnarray}
\label{eq:non_grid_bearing_loss}
L_\alpha = \frac{1}{n} \sum_{i=1}^n ||\mathbf{p_i} - \mathbf{\hat{p}_i}||_{2},
\end{eqnarray}
where $n$ is the number of sampling points; $\alpha$ is a parameter, $\alpha = \{\theta, \psi, f, k_1\}$; and $\mathbf{\hat{p}}$ is the ground-truth value of world coordinates $\mathbf{p}$. Note that $L_{\theta}$ indicates that the loss function uses a predicted $\theta$ and ground-truth parameters for $\psi$, $f$, and $k_1$. Additionally, $L_{\psi}$, $L_f$, and $L_{k_1}$ are determined in the same manner. We obtain the world coordinates $\mathbf{p}$ from the image coordinates in sampled points. The sampled points are projected from a unit sphere. For sampling on the unit sphere, we use uniform distribution within valid incident angles that depend on $k_1$. The loss function achieved stably to calibrate cameras using the unit sphere. The joint loss is defined as
\begin{eqnarray}
\label{eq:total_loss}
L = w_{\theta}L_{\theta} + w_{\psi}L_{\psi} + w_{f}L_{f} + w_{k_1}L_{k_1},
\end{eqnarray}
where $w_{\theta}$, $w_{\psi}$, $w_f$, and $w_{k_1}$ are the joint weights of $\theta$, $\psi$, $f$, and $k_1$, respectively. Although this loss function can effectively train networks, we need to determine the joint weights for each camera parameter.
Wakai~\etal{}~\cite{Wakai2021} used the joint weights set to the same values. To determine the optimal joint weights, they needed to repeat training and validation. However, they did not search for the optimal joint weights because of large computational cost.

To address this problem, we surprisingly found that numerical simulations instead of training can analyze the loss landscapes. This loss function can be divided into two steps: predicting camera parameters from a image and projecting sampled points using camera parameters. The latter step requires only the sampled points and camera parameters. Therefore, we focused on the latter step independent of input images. \Fref{fig:harmonic_loss_plot} (a) shows the loss landscapes for camera parameters along normalized camera parameters. The landscapes express that the magnitude of loss values of the focal length is the smallest among $\theta$, $\psi$, $f$, and $k_1$, and the focal length is relatively hard to train. Our investigation suggests that the optimal joint loss weights $w$ are estimated as follows: We calculate areas $S$ under the loss function $L$ for $\theta$, $\psi$, $f$, and $k_1$. Assuming practical conditions, we set the ground-truth values to $0.5$, which means that the center of the normalized parameter ranges from $0$ to $1$ in~\fref{fig:harmonic_loss_plot} (a). This area $S$ integrating $L$ for the interval $0$ to $1$ is illustrated in~\fref{fig:harmonic_loss_plot} (b) and is given by
\begin{eqnarray}
\label{eq:harmonic_weight_area}
S_{\alpha} = \int_0^1 L_\alpha d\alpha = \int_0^1 \frac{1}{n} \sum_{i=1}^n ||\mathbf{p_i} - \mathbf{\hat{p}_i}||_{2}~~ d\alpha,
\end{eqnarray}
These areas $S$ represent the magnitude of each loss for $\theta$, $\psi$, $f$, and $k_1$. Therefore, we define the joint weights $w$ in \eref{eq:total_loss} using normalization as follows:
\begin{eqnarray}
\label{eq:harmonic_normalized_weight}
w_\alpha = \tilde{w}_\alpha ~/~ W,
\end{eqnarray}
where $\tilde{w}_\alpha = 1 / S_\alpha$ and $W = \sum_\alpha \tilde{w}_{\alpha}$. We call a loss function using the weights in~\eref{eq:harmonic_normalized_weight} "harmonic non-grid bearing loss (HNGBL)." As stated above, our joint weights can alleviate the bias of the magnitude of the loss for camera parameters. Remarkably, we determine these weights without training.

\section{Experiments}
\label{sec:experiments}
To validate the adaptiveness of our method to various types of fisheye cameras, we conducted massive experiments using large-scale synthetic images and off-the-shelf fisheye cameras.

\subsection{Datasets}
\label{subsec:datasets}
\setlength{\tabcolsep}{4pt}
\begin{table}[t]
\caption{Distribution of the camera parameters for our train set}
\label{table:DatasetDistribution}
\vspace{-1mm}
\centering
\scalebox{0.85}{
\begin{tabular}{ccc}
\hline\noalign{\smallskip}
Parameters & Distribution & Range or values$^1$ \\
\noalign{\smallskip}
\hline\hline
\noalign{\smallskip}
Pan $\phi$ & Uniform & $[0, 360)$ \\
\noalign{\smallskip}
\hline
\noalign{\smallskip}
\multirow{3}{*}{Tilt $\theta$} & Mix & Normal 70\%, Uniform 30\% \\
  & Normal & $\mu=0, \sigma=15$ \\
  & Uniform & $[-90, 90]$ \\
\noalign{\smallskip}
\hline
\noalign{\smallskip}
\multirow{3}{*}{Roll $\psi$} & Mix & Normal 70\%, Uniform 30\% \\
  & Normal & $\mu=0, \sigma=15$ \\
  & Uniform & $[-90, 90]$ \\
\noalign{\smallskip}
\hline
\noalign{\smallskip}
\multirow{2}{*}{Aspect ratio} & \multirow{2}{*}{Varying} & \{1/1 9\%, 5/4 1\%, 4/3 66\%, \\& & 3/2 20\%, 16/9 4\%\} \\
\noalign{\smallskip}
\hline
\noalign{\smallskip}
Focal length $f$ & Uniform & $[6, 15]$ \\
\noalign{\smallskip}
\hline
\noalign{\smallskip}
Distortion $k_1$ & Uniform & $[-1/6, 1/3]$ \\
\noalign{\smallskip}
\hline
\noalign{\smallskip}
~~~Max angle $\eta_{\rm{max}}$~~~ & Uniform & $[84, 96]$ \\
\noalign{\smallskip}
\hline
\noalign{\smallskip}
\multicolumn{3}{l}{\scalebox{0.85}{~$^1$ Units: $\phi$, $\theta$, $\psi$, and $\eta_{\rm{max}}$ [deg]; $f$ [mm]; $k_1$ [dimensionless]}} \\
\end{tabular}
}
\end{table}
\setlength{\tabcolsep}{1.4pt}
We used two large-scale datasets of outdoor panoramas called the StreetLearn dataset (Manhattan 2019 subset)~\cite{Mirowski2019} and the SP360 dataset~\cite{Chang2018}. First, we divided each dataset into train and test sets following in~\cite{Wakai2021}: $55,599$ train and $161$ test images for StreetLearn, and $19,038$ train and $55$ test images for SP360. Second, we generated image patches, with a $224$-pixel image height ($H_{img}$) and image width ($W_{img} = H_{img} \cdot A$), where $A$ is the aspect ratio, from panorama images: $555,990$ train and $16,100$ test image patches for StreetLearn, and $571,140$ train and $16,500$ test image patches for SP360. \tref{table:DatasetDistribution} shows the random distribution of the train set when we generated image patches using camera models with the maximum incident angle $\eta_{\rm{max}}$. The test set used the uniform distribution instead of the mixed and varying distribution used for the train set. During the generation step, we set the minimum image circle diameter to the image height, assuming practical conditions.

\subsection{Off-the-shelf fisheye cameras}
\label{subsec:actual_fisheye_images}
We evaluated off-the-shelf fisheye cameras because fisheye cameras have complex lens distortion, unlike narrow-view cameras. \tref{table:comparison_of_off_the_shelf_cameras} shows various fisheye cameras that we used for evaluation. Note that we only used the front camera in the QooCam8K camera, which has both front and rear cameras. We captured outdoor fisheye images in Kyoto, Japan using the off-the-shelf cameras.

\setlength{\tabcolsep}{4pt}
\begin{table}[t]
\caption{Off-the-shelf fisheye cameras with experimental IDs}
\vspace{-1mm}
\centering
\scalebox{0.82}{
\begin{tabular}{ccc}
\hline\noalign{\smallskip}
ID & Camera body & Camera lens \\
\noalign{\smallskip}
\hline
\noalign{\smallskip}
1 & Canon EOS 6D Mark II & Canon EF8-15mm F4L Fisheye USM \\
2 & Canon EOS 6D Mark II & Canon EF15mm F2.8 Fisheye \\
\multirow{2}{*}{3} & \multirow{2}{*}{Panasonic LUMIX GM1} & Panasonic LUMIX \\
& & G FISHEYE 8mm F3.5 \\
4 & FLIR BFLY-U3-23S6C & FIT FI-40 \\
5 & FLIR FL3-U3-88S2 & FUJIFILM FE185C057HA-1 \\
6 & KanDao QooCam8K & Built-in \\
\hline
\end{tabular}
}
\label{table:comparison_of_off_the_shelf_cameras}
\end{table}
\setlength{\tabcolsep}{1.4pt}

\subsection{Parameter and network settings}
\label{subsec:parameter_and_network_settings}
To simplify the camera model, we fixed $d_u = d_v$ and the principal points $(c_u,c_v)$ as the center of the image. Because the scale factor depends on the focal length and image sensor size, which is arbitrary for undistortion, we assumed that the image sensor height was 24 mm, which corresponds to a full-size image sensor. We ignored the arbitrary translation vector $\trans$. Because the origin of the pan angle is arbitrary, we provided the pan angle for training and evaluation. Therefore, we focused on four trainable parameters, that is, tilt angle $\theta$, roll angle $\psi$, focal length $f$, and a distortion coefficient $k_1$, in our method. Note that we considered camera rotation based on the horizontal line, unlike calibration methods~\cite{Lee2020,Lee2021} under the Manhattan world assumption.

We optimized our network for a $32$ mini-batch size using a rectified Adam optimizer~\cite{Liu2019}, whose weight decay was $0.01$. We set the initial learning rate to $1\times10^{-4}$ and multiplied the learning rate by $0.1$ at the $50$th epoch. Additionally, we set the joint weights in~\eref{eq:total_loss} using $w_{\theta} = 0.103$, $w_{\psi} = 0.135$, $w_{f} = 0.626$, and $w_{k_1} = 0.136$.

\subsection{Experimental results}
\label{subsec:experimental_resuls}
\setlength{\tabcolsep}{4pt}
\begin{table}[t]
\caption{Feature summarization of the conventional methods and our method}
\label{table:comparison_of_sota}
\vspace{-1mm}
\centering
\scalebox{0.82}{
\begin{tabular}{ccccc}
\hline\noalign{\smallskip}
Method & DL$^2$ & Rot$^2$ & Dist$^2$ & Projection \\
\noalign{\smallskip}
\hline
\noalign{\smallskip}
Santana-Cedr\'{e}s~\cite{Santana2016} & & & \checkmark & Perspective \\
Liao~\cite{liao2020} & \checkmark & & \checkmark & Perspective \\
Yin~\cite{Yin2018} & \checkmark & & \checkmark & Generic camera \\
Chao~\cite{Chao2020}$^1$ & \checkmark & & \checkmark & -- \\
L\'{o}pez-Antequera~\cite{Lopez2019} & \checkmark & \checkmark & \checkmark & Perspective \\
Wakai~\cite{Wakai2021} & \checkmark & \checkmark & \checkmark & Equisolid angle  \\
Ours & \checkmark & \checkmark & \checkmark & Proposed generic camera \\
\hline
\noalign{\smallskip}
\multicolumn{5}{l}{\scalebox{0.85}{~$^1$ Using a generator for undistortion}} \\
\multicolumn{5}{l}{\scalebox{0.85}{~$^2$ DL denotes learning-based method; Rot denotes rotation; Dist denotes distortion}} \\
\end{tabular}
}
\end{table}
\setlength{\tabcolsep}{1.4pt}

\setlength{\tabcolsep}{4pt}
\begin{table*}[t]
\caption{Comparison of the absolute parameter errors and reprojection errors on the test set for our generic camera model}
\label{table:comparison_of_extrinsics}
\vspace{-1mm}
\centering
\scalebox{0.85}{
\begin{tabular}{ccccccccccc}
\hline\noalign{\smallskip}
 & \multicolumn{5}{c}{StreetLearn} & \multicolumn{5}{c}{SP360} \\
\cmidrule(lr){2-6} \cmidrule(lr){7-11}
Method & \multicolumn{4}{c}{Mean absolute error $\downarrow$} & REPE $\downarrow$ & \multicolumn{4}{c}{Mean absolute error $\downarrow$} & REPE $\downarrow$ \\
& Tilt $\theta$ [deg] & Roll $\psi$ [deg] & $f$ [mm] & $k_1$ & [pixel] & Tilt $\theta$ [mm] & Roll $\psi$ [deg] & $f$ [mm] & $k_1$ & [pixel] \\
\noalign{\smallskip}
\hline
\noalign{\smallskip}
L\'{o}pez-Antequera~\cite{Lopez2019} & 27.60 & 44.90 & 2.32 & -- & 81.99 & 28.66 & 44.45 & 3.26 & -- & 84.56 \\
Wakai~\cite{Wakai2021} & 10.70 & 14.97 & 2.73 & -- & 30.02 & 11.12 & 17.70 & 2.67 & -- & 32.01 \\
\hline
\noalign{\smallskip}
Ours w/o HNGBL$^1$ & \phantom{0}7.23 & \phantom{0}7.73 & 0.48 & 0.025 & 12.65 & \phantom{0}6.91 & \phantom{0}8.61 & 0.49 & 0.030 & 12.57 \\
Ours & \textbf{\phantom{0}4.13} & \textbf{\phantom{0}5.21} & \textbf{0.34} & \textbf{0.021} & \textbf{\phantom{0}7.39} & \textbf{\phantom{0}3.75} & \textbf{\phantom{0}5.19} & \textbf{0.39} & \textbf{0.023} & \textbf{\phantom{0}7.39}\\
\hline
\noalign{\smallskip}
\multicolumn{11}{l}{\scalebox{0.85}{~$^1$ "Ours w/o HNGBL" refers to replacing HNGBL with non-grid bearing loss~\cite{Wakai2021}}} \\
\end{tabular}
}
\end{table*}
\setlength{\tabcolsep}{1.4pt}
In~\tref{table:comparison_of_sota}, we summarize the features of the conventional methods. We implemented the methods according to the implementation details provided in the corresponding papers, with the exception that StreetLearn and SP360 were used for training the methods of Chao~\cite{Chao2020}, L\'{o}pez-Antequera~\cite{Lopez2019}, and Wakai~\cite{Wakai2021}. For the method of Santana-Cedr\'{e}s~\cite{Santana2016}, we excluded test images with few lines because this method requires many lines for calibration.

\subsubsection{Parameter and reprojection errors}
\label{subsubsec:parameter_and_reprojection_errors}
To validate the accuracy of the predicted camera parameters, we compared methods that can predict rotation and distortion parameters. We evaluated the mean absolute errors of the camera parameters and mean reprojection errors (REPE) on the test set for our generic camera model. \Tref{table:comparison_of_extrinsics} shows that our method achieved the lowest mean absolute errors and REPE among all methods. This REPE reflected the errors of both extrinsic and intrinsic parameters. To calculate the REPE, we generated $32,400$ uniform world coordinates on the unit sphere within less than $90^{\circ}$ because of the lack of calibration points for the image-based calibration methods. L\'{o}pez-Antequera's method~\cite{Lopez2019} did not seem to work well because it expects non-fisheye input images. Our method substantially reduced focal length errors and camera rotation errors (tilt and roll angles) by $86$\% and $66$\%, respectively, on average for the two datasets compared with Wakai's method~\cite{Wakai2021}. Furthermore, our method reduces the REPE by $76$\% on average for the two datasets compared with Wakai's method~\cite{Wakai2021}. Therefore, our method predicted accurate extrinsic and intrinsic camera parameters.

We also evaluated our method, referred to as "Ours w/o HNGBL," replacing our loss function with non-grid bearing loss~\cite{Wakai2021} to analyze the performance of our loss function, as shown in~\tref{table:comparison_of_extrinsics}. This result demonstrates that our loss function effectively reduced the rotation errors in the tilt and roll angles by $3.05^\circ$ on average for the two datasets compared with the "Ours w/o HNGBL" case. In addition to rotation errors, the REPE for our method with HNGBL was $5.22$ pixels on average for the two datasets smaller than that for "Ours w/o HNGBL." These results suggest that our loss function enabled networks to accurately predict not only focal length but also other camera parameters.

\subsubsection{Comparison using PSNR and SSIM}
\label{subsubsec:comparison_using_psnr_and_ssim}
\setlength{\tabcolsep}{4pt}
\begin{table}[t]
\caption{Comparison of mean PSNR and SSIM on the test set for our generic camera model}
\label{table:comparison_of_psnr_ssim}
\vspace{-1mm}
\begin{threeparttable}
\centering
\scalebox{0.85}{
\begin{tabular}{ccccc}
\hline\noalign{\smallskip}
\multirow{2}{*}{Method} & \multicolumn{2}{c}{StreetLearn} & \multicolumn{2}{c}{SP360} \\
\cmidrule(lr){2-3} \cmidrule(lr){4-5}
 & PSNR $\uparrow$ & SSIM $\uparrow$ & PSNR $\uparrow$ & SSIM $\uparrow$ \\
\noalign{\smallskip}
\hline
\noalign{\smallskip}
Santana-Cedr\'{e}s~\cite{Santana2016} & 14.65 & 0.341 & 14.26 & 0.390  \\
Liao~\cite{liao2020} & 13.71 & 0.362 & 13.85 & 0.404 \\
Yin~\cite{Yin2018} & 13.91 & 0.349 & 14.03 & 0.390 \\
Chao~\cite{Chao2020} & 16.13 & 0.409 & 15.88 & 0.449 \\
~~~~L\'{o}pez-Antequera~\cite{Lopez2019}~~~~ & 17.88 & 0.499 & 16.24 & 0.486 \\
Wakai~\cite{Wakai2021} & 21.57 & 0.622 & 20.98 & 0.639 \\
\hline
\noalign{\smallskip}
Ours w/o HNGBL\tnote{1} & 27.41 & 0.801 & 26.49 & 0.801 \\
Ours & \textbf{29.01} & \textbf{0.838} & \textbf{28.10} & \textbf{0.835} \\
\hline
\end{tabular}
}
\begin{tablenotes}\footnotesize
\item[1] \scalebox{0.85}{"Ours w/o HNGBL" refers to replacing HNGBL with non-grid bearing loss~\cite{Wakai2021}}
\end{tablenotes}
\end{threeparttable}
\end{table}
\setlength{\tabcolsep}{1.4pt}

\setlength{\tabcolsep}{4pt}
\begin{table*}[t!]
\caption{Comparison of mean PSNR on the test set for the trigonometric function models}
\label{table:comparison_of_trigonometric_psnr}
\vspace{-1mm}
\centering
\scalebox{0.81}{
\begin{tabular}{ccccccccccc}
\hline\noalign{\smallskip}
 & \multicolumn{5}{c}{StreetLearn} & \multicolumn{5}{c}{SP360} \\
\cmidrule(lr){2-6} \cmidrule(lr){7-11}
Method & Stereo- & Equi- & Equisolid & \phantom{0}Ortho-\phantom{0} & \multirow{2}{*}{All} & Stereo- & Equi- & Equisolid & \phantom{0}Ortho-\phantom{0} & \multirow{2}{*}{All} \\
& graphic & distance & angle & gonal & & graphic & distance & angle & gonal & \\
\noalign{\smallskip}
\hline
\noalign{\smallskip}
Santana-Cedr\'{e}s~\cite{Santana2016} & 14.68 & 13.20 & 12.49 & 10.29 & 12.66 & 14.25 & 12.57 & 11.77 & \phantom{0}9.34 & 11.98 \\
Liao~\cite{liao2020} & 13.63 & 13.53 & 13.52 & 13.74 & 13.60 & 13.76 & 13.66 & 13.67 & 13.92 & 13.75 \\
Yin~\cite{Yin2018} & 13.81 & 13.62 & 13.59 & 13.77 & 13.70 & 13.92 & 13.74 & 13.72 & 13.94 & 13.83 \\
Chao~\cite{Chao2020} & 15.86 & 15.12 & 14.87 & 14.52 & 15.09 & 15.60 & 15.02 & 14.83 & 14.69 & 15.03 \\
L\'{o}pez-Antequera~\cite{Lopez2019} & 17.84 & 16.84 & 16.43 & 15.15 & 16.57 & 15.72 & 14.94 & 14.68 & 14.52 & 14.97 \\
Wakai~\cite{Wakai2021} & 22.39 & 23.62 & 22.91 & 17.79 & 21.68 & 22.29 & 22.65 & 21.79 & 17.54 & 21.07 \\
\hline
\noalign{\smallskip}
Ours w/o HNGBL$^1$ & 26.49 & 29.08 & 28.56 & \textbf{23.97} & 27.02 & 25.35 & 28.53 & 28.26 & 23.85 & 26.50 \\
Ours & \textbf{26.84} & \textbf{30.10} & \textbf{29.69} & 23.70 & \textbf{27.58} & \textbf{25.74} & \textbf{29.28} & \textbf{28.95} & \textbf{23.93} & \textbf{26.98} \\
\hline
\noalign{\smallskip}
\multicolumn{11}{l}{\scalebox{0.85}{~$^1$ "Ours w/o HNGBL" refers to replacing HNGBL with non-grid bearing loss~\cite{Wakai2021}}} \\
\end{tabular}
}
\end{table*}
\setlength{\tabcolsep}{1.4pt}
To demonstrate validity and effectiveness in images, we used the peak signal-to-noise ratio (PSNR) and structural similarity (SSIM)~\cite{Wang2004} for intrinsic parameters. When performing undistortion, extrinsic camera parameters are arbitrary because we consider only intrinsic camera parameters, image coordinates, and incident angles. \Tref{table:comparison_of_psnr_ssim} shows the performance of undistortion on the test set for our generic camera model. Our method notably improved the image quality of undistortion by $7.28$ for the PSNR and $0.206$ for the SSIM on average for the two datasets compared with Wakai's method~\cite{Wakai2021}.

To validate the dependency of the four types of fisheye camera models, we also evaluated the performance on the trigonometric function models in~\tref{table:comparison_of_trigonometric_psnr}. Although orthogonal projection decreased PSNR, our method addressed all the trigonometric function models; hence, our method had the highest PSNR in all cases. This suggests that our generic camera model precisely behaved like a trigonometric function model. Therefore, our method has the potential to calibrate images from various fisheye cameras.

\subsubsection{Qualitative evaluation}
\label{sec:qualitative_evaluation}
\begin{figure*}[t!]
\centering
\includegraphics[width=0.83\hsize]{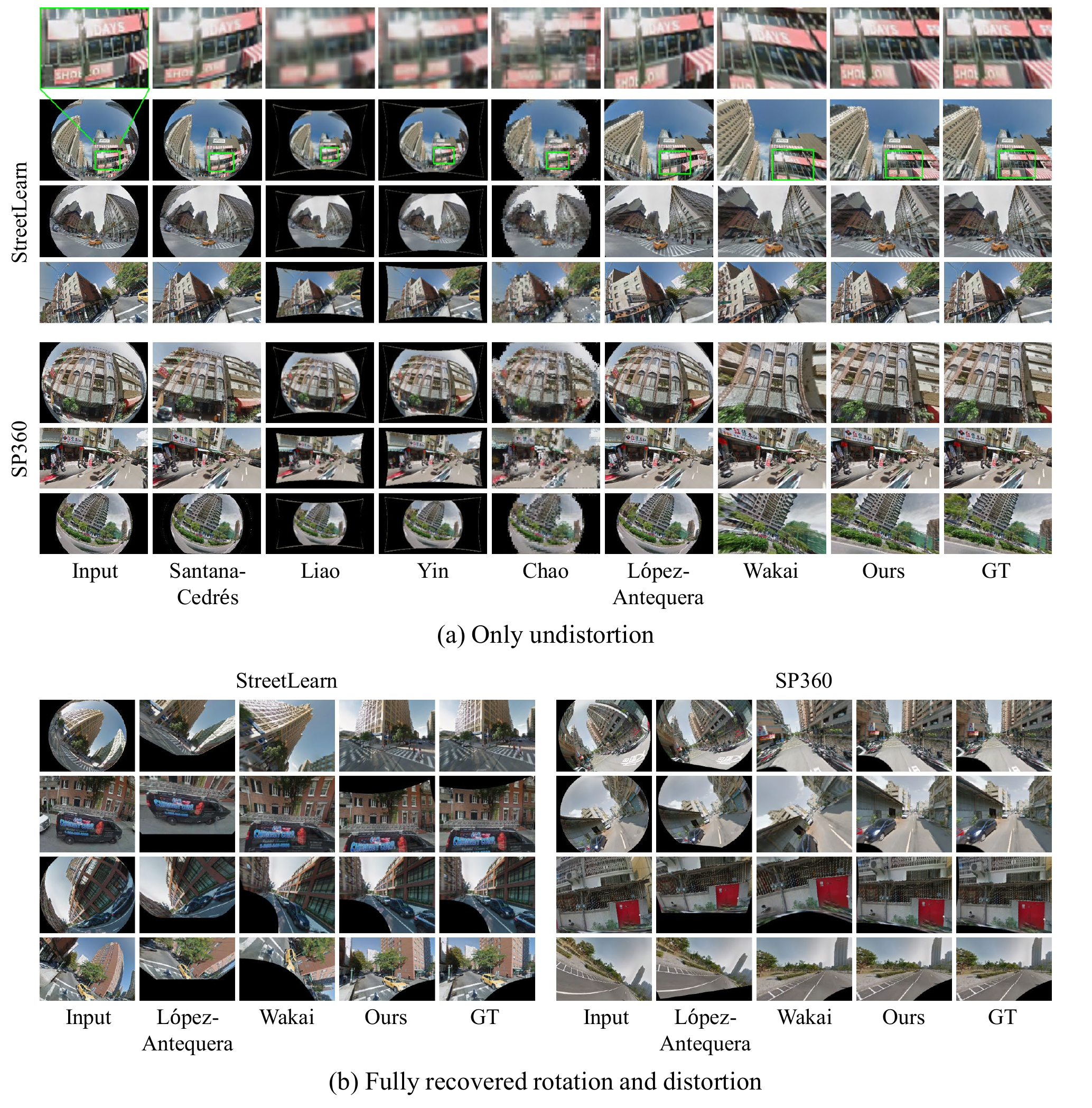}
\caption{Qualitative results on the test images for our generic camera model. (a) Undistortion results shown in the input image, results of the compared methods (Santana-Cedr\'{e}s~\cite{Santana2016}, Liao~\cite{liao2020}, Yin~\cite{Yin2018}, Chao~\cite{Chao2020}, L\'{o}pez-Antequera~\cite{Lopez2019}, and Wakai~\cite{Wakai2021}), our method, and the ground-truth image from left to right. (b) Fully recovered rotation and distortion shown in the input image, results of the compared methods (L\'{o}pez-Antequera~\cite{Lopez2019} and Wakai~\cite{Wakai2021}), our method, and the ground-truth image from left to right.}
\label{fig:compare_undist_rotate_synth}
\end{figure*}

\begin{figure*}[t]
\centering
\includegraphics[width=0.99\hsize]{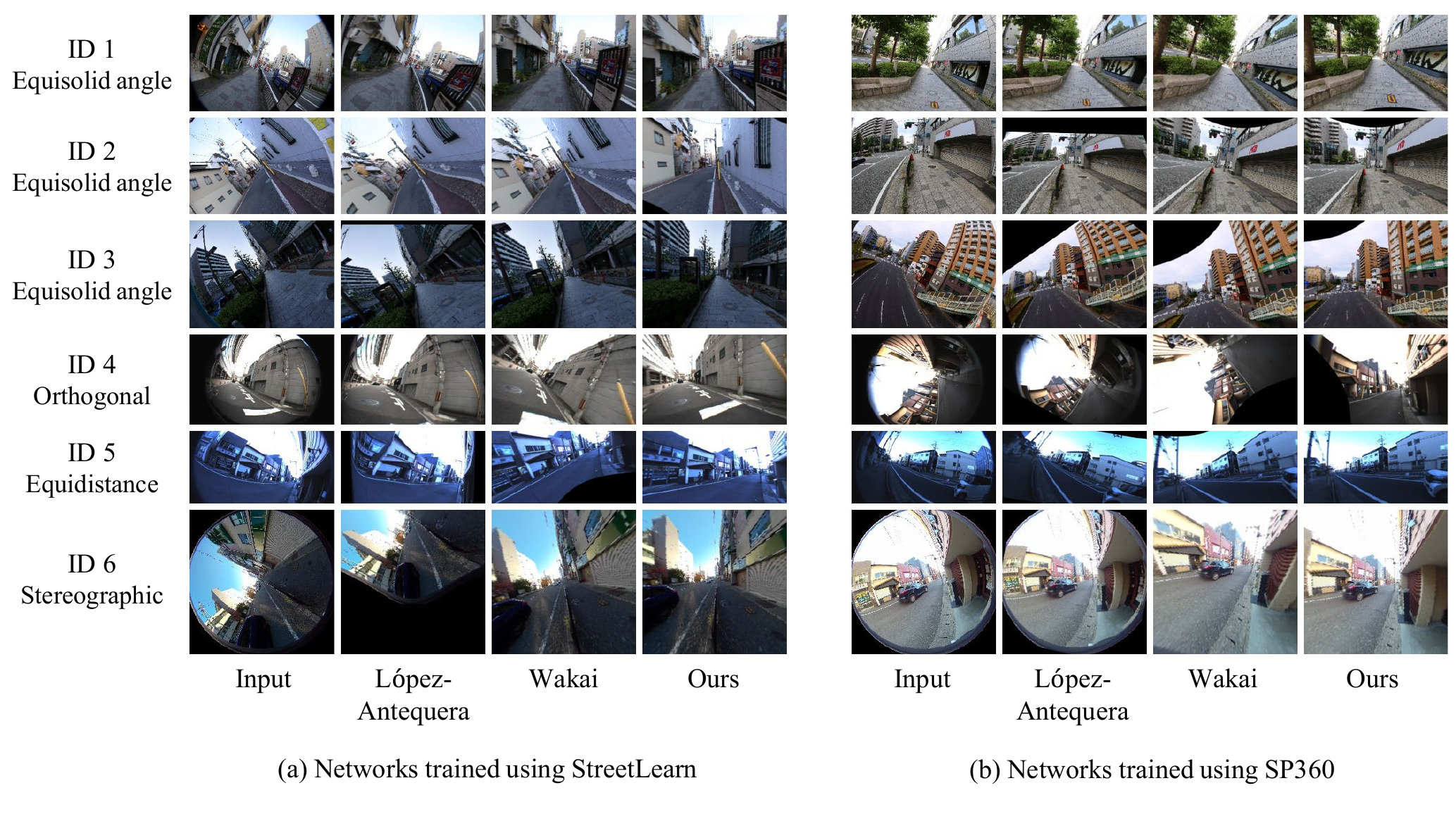}
\caption{Qualitative results of fully recovering rotation and fisheye distortion for the off-the-shelf cameras shown in the input image, results of the compared methods (L\'{o}pez-Antequera~\cite{Lopez2019} and Wakai~\cite{Wakai2021}), and our method from left to right for each image. The IDs correspond to IDs in~\tref{table:comparison_of_off_the_shelf_cameras}, and the projection names are attached to the IDs from specifications (ID: 3--5) and our estimation (ID: 1, 2, and 6). Qualitative results of the methods trained using StreetLearn~\cite{Mirowski2019} and SP360~\cite{Chang2018} datasets as shown in (a) and (b), respectively.}
\label{fig:compare_image_actual_rotate}
\end{figure*}
We evaluated the performance of undistortion and full recovery not only for synthetic images but also off-the-shelf cameras to describe the image quality after calibration.

\textbf{Synthetic images:} \Fref{fig:compare_undist_rotate_synth} shows the qualitative results on the test set for our generic camera model. Our results are the most similar to the ground-truth images in terms of undistortion, and fully recovering rotation and fisheye distortion. Our method worked well for various types of distortion and scaling. By contrast, it was difficult to calibrate circumferential fisheye images with large distortion using Santana-Cedr\'{e}s's method~\cite{Santana2016}, Liao's method~\cite{liao2020}, Yin's method~\cite{Yin2018}, and Chao's method~\cite{Chao2020}. Furthermore, L\'{o}pez-Antequera's~\cite{Lopez2019} and Wakai's~\cite{Wakai2021} methods did not remove distortion, although the scale was close to the ground truth.

When fully recovering rotation and distortion, L\'{o}pez-Antequera's~\cite{Lopez2019} and Wakai's~\cite{Wakai2021} methods tended to predict camera rotation with large errors in the tilt and roll angles. As shown in~\fref{fig:compare_undist_rotate_synth}, our synthetic images consisted of zoom-in images of parts of buildings and zoom-out images of skyscrapers. Our method processed both types of images, that is, it demonstrated scale robustness.

\textbf{Off-the-shelf cameras:} We also validated calibration methods using off-the-shelf fisheye cameras to analyze the performance of actual complex fisheye distortion. Note that studies on the conventional learning-based methods in~\tref{table:comparison_of_sota} reported evaluation results using only synthetic fisheye images. \Fref{fig:compare_image_actual_rotate} shows the qualitative results of fully recovering rotation and fisheye distortion for methods that predicted intrinsic and extrinsic camera parameters. These methods were trained using the StreetLearn~\cite{Mirowski2019} or SP360~\cite{Chang2018} datasets. The results for L\'{o}pez-Antequera's method had distortion and/or rotation errors. Our method outperformed Wakai's method~\cite{Wakai2021}, which often recovered only distortion for all our cameras. Our fully recovered images demonstrated the effectiveness of our method for off-the-shelf fisheye cameras with various types of projection.

In all the calibration methods, images captured by off-the-shelf cameras seemingly degraded the overall performance in the qualitative results compared with synthetic images. This degradation probably occurred because of the complex distortion of off-the-shelf fisheye cameras and the dataset domain mismatch between the two panorama datasets and our captured images. Overall, our method outperformed the conventional methods in the qualitative evaluation of off-the-shelf cameras. As described above, our method precisely recovered both rotation and fisheye distortion using our generic camera model.

\section{Conclusion}
We proposed a learning-based calibration method using a new generic camera model to address various types of camera projection. Additionally, we introduced a novel loss function that has optimal joint weights determined without training. These weights can alleviate the bias of the magnitude of each loss for four camera parameters. As a result, we enabled networks to precisely predict both extrinsic and intrinsic camera parameters. Extensive experiments demonstrated that our proposed method substantially outperformed conventional geometric-based and learning-based methods on two large-scale datasets. Moreover, we demonstrated that our method fully recovered rotation and distortion using various off-the-shelf fisheye cameras. To improve the calibration performance in off-the-shelf cameras, in future work, we will study the dataset domain mismatch.

{\small
\bibliographystyle{ieee_fullname}
\bibliography{egbib}
}

\end{document}